# Urdu-English Machine Transliteration using Neural Networks

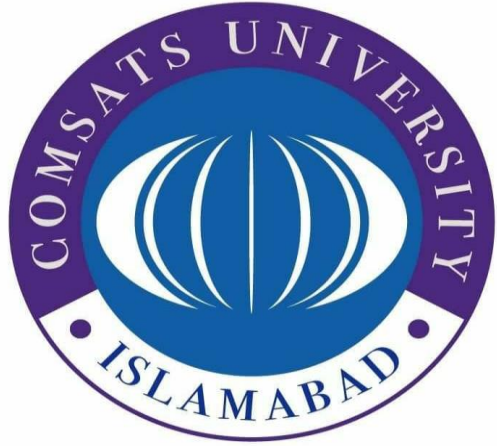

*By*

*Usman Mohy ud Din*

*CIIT/FA16-RCS-007/LHR*

MS Thesis

In

**MS (CS)**

# COMSATS University Islamabad, Lahore Campus-Pakistan

Spring, 2019

# DEDICATION

I would like to dedicate this thesis to my loving family. Their prayers, support, encouragement and love have sustained me throughout my life.

# ACKNOWLEDGEMENTS

*"Recite in the Name of Your Lord Who created. He created the human being from a clot. Recite and your Lord is the Most Honourable, who taught (to write) with the pen, taught the human being what he knew not..."* Holy Quran (Alaq 96: 1-5)

First of all, I would like to thanks to Almighty Allah for giving me strength and courage to step forward for this work. I would like to thanks my supervisor Dr. Muhammad Waqas Anwar who guided me in right direction and encouraged me throughout my research work. He granted me with his encouragement, guidance and good ideas throughout my research work. He kept me motivated for in time completion of this research study. He was always there to support me when I was struggling to make a balance in work-life and studies. I would have been completely lost without his support.

In Soliton technologies, I gratefully thank to Imran Ali *(ground-manager)*, Bilal khan, Meera, Dr. Zunaira, Giri Raj, Adil Saeed, Hassan Hayat (*scrum-masters*) and all my co-workers (skipping the names because list is too long) with whom I worked in Quality, DevOps and validation teams during my MS. The way they supported me, celebrated each little success of academics, compromised on my work-schedule so that I can concentrate on studies, is really heart whelming.

In Comsats, I am thankful to my fellows Muhammad Awais, Noor ullah Khan, Mehmood, Ammar, Hafiz Faraz, Shahzad and Romila Aziz for their help and support. I am also thankful to Waseem khan who is key person behind all the troubleshooting and resolving all infrastructures and network related issues during my studies. I am indebted to Kanza Hamid for her valuable guidance, suggestions and reviews in order to improve the quality of my work.

In NTU, I am humbled and honoured to meet Dr. Asif and Abdul basit who provided the hardware to train the system. Without their support, this would not be possible.

Last but not the least, I am thankful to my QAU friends who have been in my life. Their unique way to motivate others is always pleasurable. Their help and advices during rough and tough patches were valuable.

**Usman Mohy ud Din**

**CIIT/FA16-RCS-007/LHR**

# ABSTRACT

## Urdu-English Machine Transliteration using Neural Networks

Machine translation has gained much attention in recent years. It is a sub-field of computational linguistic which focus on translating text from one language to other language. Among different translation techniques, neural network currently leading the domain with its capabilities of providing a single large neural network with attention mechanism, sequence-to-sequence and long-short term modelling. Despite significant progress in domain of machine translation, translation of out-of-vocabulary words(OOV) which include technical terms, named-entities, foreign words are still a challenge for current state-of-art translation systems, and this situation becomes even worse while translating between low resource languages or languages having different structures. Due to morphological richness of a language, a word may have different meninges in different context. In such scenarios, translation of word is not only enough in order provide the correct/quality translation. Transliteration is a way to consider the context of word/sentence during translation. For low resource language like Urdu, it is very difficult to have/find parallel corpus for transliteration which is large enough to train the system. In this work, we presented transliteration technique based on Expectation Maximization (EM) which is un-supervised and language independent. Systems learns the pattern and out-of-vocabulary (OOV) words from parallel corpus and there is no need to train it on transliteration corpus explicitly. This approach is tested on three models of statistical machine translation (SMT) which include *phrase-based*, *hierarchical phrase-based* and *factor based* models and two models of neural machine translation which include *LSTM* and *transformer* model. On SMT models, there is gain of 0.63 to 0.91 in BLEU score while on NMT models, there is gain of 1.28 to 2.05 in BLEU which are better than previous baseline scores. Our approach shows promising results in translation of Urdu text into English which is mostly neglected due to its complexities. We also discussed the results, different challenges faced during this work and effect of right pre-processing techniques.

# Table of Contents

# LIST OF FIGURES

# LIST OF TABLES

**Chapter 1**
**Introduction**

## 1.1    Introduction

Researchers of computer science are interested in developing systems to improve the interaction between humans and computers [1]. Natural Language System represent the most important field of investigation to serve this interest [2]. Language is one of the most powerful tools of any living being to convey their thoughts to the other but it is only possible if the communicating subjects have the same language. A language can be expressed as a series of spoken sounds and words or gestures i.e. body language. Every species has its own language e.g. animals, birds produce particular sounds to communicate with their species fellows. Similarly, humans use a group of words collectively called language for communication. On the present time according to the survey about 65,000 languages are spoken in the worldwide. Every human has its own native language according to their culture and region if two persons of the same region and culture want to communicate they would use their native language. But if people belong to different regions or culture then they might have different languages which can be so different for them. If they want to communicate then they must understand other's language or there must be some middleman or translator who can translate their words for them but even for a middleman to master or learn all or most of the existing languages is impossible. The present era is the time of advanced technologies and artificially intelligent agents. In this era where most of our daily chores are performed by machines of artificially intelligent systems so why not machines should work as a middleman translator for different people. Because for a machine it's not difficult to learn 65,000 languages and it has already started and emerged as a complete field by name natural language processing and has progressed a lot.

## 1.2 Urdu Language

Urdu is a free-order language which belongs to the Indo-Aryan family of languages. Urdu is developed under the influence of Arabic, Persian, Turkish and Hindi. The word "Urdu" itself is derived from Turkish. Urdu is the national language of Pakistan and one of official language in India and Jammu Kashmir. It is a popular language in other South Asian countries like Bangladesh and Afghanistan. Urdu has almost 104 million speakers around the globe. Urdu character-set with 38 characters is the super set of Arabic and Persian character-sets with 28 and 32 characters respectively. Usually, characters of Urdu language have four shapes, 1) Isolated, 2) as the First letter of the word, 3) Middle and 4) as the Last letter of the word. Urdu has some interesting characteristics as compared to other languages. Urdu follows SOV (Subject-Object-Verb) structure in a simple sentence. In Urdu, the text is written in right to left manner while digits in sentences are written in left to right. Urdu has additional tenses as compared to English. In Urdu, for example, the past indefinite tense of English language has three sub-types near past, absolute past and distant past which will discuss further in upcoming chapters. There is gender associated with Urdu words like adjectives and verbs. For example, in Urdu literature, table is considered as masculine while chair is considered as feminine. Moreover, verb changes its shape w.r.t subject's gender (male or female) and number (singular or plural).

## 1.3   Machine Translation

Machine Translation is an important filed in Natural Language Processing. With invent of web 2.0, online content is increasing with rapid growth. It is need of hour to have such systems which can translate text from one language to other to make it useful for everyone.

## 1.4   Machine Translation Techniques

Machine translation can be categorized as [3]:

➢ Example Based Machine Translation
➢ Statistical Machine Translation
➢ Rule based Machine Translation
➢ Neural Machine Translation
➢ Hybrid Machine Translation

### 1.4.1   Example Based Machine Translation

Example based machine translation works on decomposing/fragmentation of source sentence, translating these fragments into target language and then re-composing those translated fragments into long sentence [4].

### 1.4.2   Statistical Machine Translation

Statistical machine translation uses statistical concepts of probabilities while translating from input to output. While translating a particular word of source sentence, if there are multiple translation exists in target language then the system computes probabilities and choose the translation which has higher probability [5]. It produces less quality translation for the pair of languages having different structures [6]. Statistical machine translation has different models that includes [7] [8]:

- ➢ Phrase based Model
- ➢ Hierarchical Phrase based
- ➢ Factor based model
- ➢ String-to-tree model
- ➢ Tree-to-string model
- ➢ Tree-to-tree model

### 1.4.3 Rule Based Machine Translation

Rule based machine translation uses hand written linguistic rules for both languages in its translation process. It requires a lot of human effort to define the rules and modifications of rules usually costs very high [9]. It has three different types:

- ➢ Direct
- ➢ Interlingua
- ➢ Transfer based

In direct machine translation, source language is directly converted to target language without using intermediate steps. In Interlingua machine translation, there are intermediate steps which contains all necessary information for generating texts of target language. Interlingua steps usually design with the intention to make it universal for all pair of languages. In transfer based translation, there is bilingual representation of both languages in intermediate steps. This intermediate steps are language dependent [10].

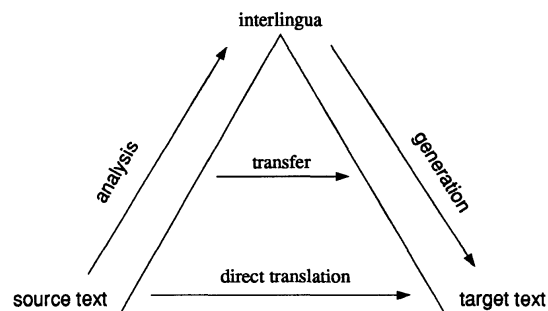

*Figure 1: Types of Rule based Machine Translation[1]*

---

### 1.4.4  Neural Machine Translation

Neural Machine Translation works by building a single large neural network which can be optimized to increase performance. In recent years, Neural Machine Translation has emerged as very popular trend in machine translation systems. According to [11], only one neural machine translation system was submitted in Conference on Machine Translation (WMT) in 2015. However, in 2017 all the submitted systems were based on neural machine translation.

### 1.4.5  Hybrid Machine Translation

In hybrid machine translation, a combination of two or more machine translation techniques is used to overcome the limitations of each technique and to enhance the quality of translation.

## 1.5  Machine Transliteration

The process of translating source language into target language or target into source in a way that context is preserved called transliteration. Transliteration becomes vital in cross lingual and information retrieval applications. There are two basic transliteration techniques, forward transliteration or mapping source language phoneme or grapheme to their equivalent target language's phoneme or grapheme and backward transliteration technique or mapping target language phoneme or grapheme to their equivalent source language phoneme or grapheme. In forward transliteration, we may have multiple translations which all are valid, but in backward transliteration there is only one valid translation exists.

### 1.5.1  Grapheme based Machine Transliteration

Grapheme is defined as the smallest unit in any writing system which may or may not carry meanings. In grapheme based approach, grapheme of source script is directly transliterated to grapheme of target script. This direct approach has four models [12]:

- ➢ Maximum Entropy Model
- ➢ Decision Tree Model
- ➢ CRF Models
- ➢ Source Channel Model

### 1.5.2  Phoneme Based Machine Transliteration

Phoneme is defined as the distinguishable unit of language. In phoneme based approach, grapheme of source script is first converted into phoneme of source script and then that phoneme is transliterated into grapheme of target script. This indirect approach has two models [12]:

➢ EMW model
➢ WEST model

### 1.5.3 Hybrid Approach

Hybrid approach uses the combination of both grapheme and phoneme based approach.

## 1.6 Translation of Nouns

Among all the parts of speech, nouns play an important role in understanding of text. Proper nouns are one of the fundamentals of any language which represents person names, geographical or organization entities. The simplest definition listed on Wikipedia states as "a name used for an individual person, place, or organization, spelled with an initial capital letter, e.g. *Jane*, *London*, and *Oxfam*". Translation of proper nouns is not as uniform as translation of other part of speech entities. Majority of languages have influence of cultural heritage of the area in which they are spoken [13]. Despite the advancements of machine translation methods, there is still room for improvement in case of machine translation of proper nouns. Following table shows the translation quality of Google[2] and Bing Microsoft[3] Translators on few Urdu sentences.

*Table I: Deficiencies in current MT systems*

| Sentence | Google Translation | Bing Translation |
|---|---|---|
| عزیز پاکستان کا وزیراعظم ہے | Dear Pakistan is prime minister | Aziz is the Prime Minister of Pakistan |
| میرا نام عظیم ہے | My name is great | My name is great |
| میرا نام عمر ہے | My name is age | My name is Umar |
| لاہور میں عمر رہتا ہے | Lives in Lahore | Lahore is life |
| لاہور ایک خوبصورت شہر ہے | There is still a beautiful city | Lahore is a beautiful city |
| قائد اعظم کا مزار کراچی میں ہے | The leader of Qayyad is in Karachi | Quaid-e-Azam Mazar is in Karachi |
| بانگ درا علامہ اقبال کی تصنیف ہے | Bangkok's portrait of Allama Iqbal | The call of the marching Bell Allama Iqbal is will be |
| صدر میں گولیاں چل رہی ہیں | The tablets are going on in the president | President are running in |

**Chapter 2**

**Literature Review**

## 2.1    Background

Janhavi and Ankit [14] surveyed deep learning techniques used in machine translation. Feed forward neural networks (FNN), recurrent neural networks (RNN), recursive auto-encoder (RAE), recursive neural networks and convolutional neural networks (CNN) are the five popular neural networks used in research. Encoder-decoder model in NMT was introduced to solve the problem of sequence-to-sequence learning. The encoder encodes the input sentence into fixed-length vector called context vector and decoder is responsible for stepping through output time steps to read from context vector. To solve the problem of encoder-decoder model for long sentences and less-similar languages, attention mechanism was introduced in neural networks. According to this study, RNN and RAE performs better among all types of neural networks. Sharmin and Pitamber [15] demonstrated the qualitative evolution of Google's SMT and NMT systems on English-Urdu language pair. The performance of both systems were compared on the statistical measure called Kappa, WER (word error rate) and sentence error rate (SER).

Chen, Firat and Bapna [16] explored the modelling and training techniques of recurrent neural machine translation (RNMT) and extended its functionality (called RNMT+) by introducing 6 bi-directional LSTM layers in encoder and 8 uni-directional layers in decoder. For the quality of translation and stability of training process, attention mechanism was also fed to softmax. Dropout, Label Smoothing and Weight Decay were used as regularization techniques in training process. They also combine the existing seq2seq models to strengthen their capabilities and devise hybrid models. Their proposed models outperform in terms of BLEU scores as compared to existing models on English –French and English – German language pairs. Yesir and Kevin [17] proposed approach used an algorithm based on sound and spellings mappings using finite state machines. The spelling based model directly maps English characters sequence to Arabic characters' sequence. In phonetic model, each English phoneme is mapped to Arabic letter sequence using P(a|e). The model's transliterated names are measured with manually translated names to measure accuracy. This spelling based mappings have higher accuracy than state-of-art phonetic based machine transliteration. The accuracy/result are based on exact matching.

Amir H. Jadinejad [18] proposed character based encoder-decoder model which consists of two recurrent neural networks and attention based mechanism. The proposed model is inspired by concept of sequence to sequence learning. Encoder consists of bidirectional neural network which convert converts input sequence into fixed length vector representation. The decoder in attention based recurrent neural network generate output sequence from that vectors. All modules of model (encoder, decoder, and attention based mechanism) are jointly trained to maximize probability. Abbas and Madiha [19] worked on English to Urdu transliteration system. In their process, English text is converted to Urdu using pronunciation and mapping rules. For English pronunciation, an arpabet based lexicon is used which is based on American accent. Inconsistencies produced by American accent was removed by applying syllabification. Frequent words were transliterated manually and out of vocabulary words problem is removed using probabilistic mapping between pronunciation and alignment.

Barret, Deniz, Jonathan and Kevin [20] presented a transfer learning method for neural machine translation The proposed method consists of two models, parent model which consists of high resource language and the child model which consists of low resource language. The parent model usually used for training purpose, after training the parent model transfers few parametrs to child model which are used for initializtion and constraints training. They claims significance improvement over encoder-decodel model. [21] used translation rules and ANN with feedforward back propagation model for English to Urdu machine translation. Maximum Entropy based tagger[4] and probabilistic natural language parser[5] are used for tagging and parsing of sentences. Information related to each word present in sentence is extracted and sentences are transformed into knowledgeable objects. Thus training data consists of grammar rules and bilingual dictionary with associated knowledge. Encoder-decoder module is used to convert training data into numeric format which is passed to ANN along with grammatical structure and attributes. For training of ANN, Levenberg-Marquardit algorithm is used with mean squared error rate of $10^{-8}$. Output of ANN is decoded into Urdu using grammatical structure and linguistic rules. BLEU, METEOR, F-Measure, Precision and Recall are used for evaluation of this system which performs well on the data that was part of training. Unseen words or those that were not part of training data, printed as it is in capitals during translation.

---

[4] https://nlp.stanford.edu/software/tagger.shtml
[5] https://nlp.stanford.edu/software/lex-parser.shtml

[22] argued that hierarchical phrase based works well as compared to classical phrase based in English to Urdu statistical machine translation. They compared their results on three different independent test sets including parallel corpus that consists of 79000 documents collected from five different sources (EMILLE, IPC[6], Quran[7], Penn Treebank[8], Afrl), monolingual data[9] and official tests that includes NIST 2008 Open Machine Translation (OpenMT) Evaluation[10], IPC and CLE[11]. They evaluate quality of translation both automatically using BLEU score and manually by ranking output using QuickJudge[12]. Because of few similarities in Urdu and Hindi, [23] tested English to Hindi translation system for English, Urdu language pair. System was based on pseudo interlingua rule based approach where Hindi language acted as an interlingua with Hindi-Urdu mapping table to generate final output. Low BLEU score observed due to gender mismatch, verb forms and differences in phonetics of Hindi and Urdu languages.

[24] compared the performance of three online machine translation systems that includes Google[13], Babylon[14] and Bing[15] on Urdu to Arabic translation. 159 parallel sentences of categories: declarative, exclamatory and imperative are used in evaluation of systems. After evaluating the performance using BLEU, METEOR and NIST, they concluded that Google translator produced better results. [25] suggested human and automatic evaluation of English to Urdu machine translation. Subjective evaluation (by humans) and objective evaluation (using BLEU, GTM, METEOR, ATEC) of Google, Babylon and Ijunoon[16] were performed at sentence level. They claimed that results of METEOR were highly correlated with parameter based human judgment. [26]

---

proposed expert system based machine translation for English to Urdu. Their system included expert system along with knowledge base and rules.

[4] presented example based machine translation for English to Urdu. First, system performed fragmentation of source sentence using idioms, connecting words and cutter points, then look for translation of each fragment in target language's corpus. The most relevant translation of each fragment was selected using Levenshtein and semantic distance algorithm. To resolve word order issue in phrases, they allowed user to customize the translation according to their needs. The proposed approach successfully handled idioms, homographs and words having sense of gender. [27] proposed transfer based approach for English to Urdu machine translation. Algorithmic structural transformation applied to handle difference in grammatical structure of both languages. Lexical parser was used to obtain parse trees of source language. Paninian theory efficiently handled case phrases and verb post positions in translation of target language. [28] explored three approaches in Urdu machine translation that are rule based machine translation (RBMT), Example based machine translation (EBMT) and statistical machine translation (SMT). They concluded that EBMT produced better BLEU score among all systems, RBMT worked well for languages having similar structure like Hindi-Urdu and SMT outperformed where linguistic resources such as annotated data are available.

[29] attempted English to Urdu statistical machine translation with development of parallel corpus. They discussed the sentence alignment issues, punctuation issues, mismatch of colons and translation issues faced during development of parallel corpora. BLEU score was used to highlight the effect of tuning in machine translation. In machine translation, different types of translation ambiguities such as lexical ambiguity, polysemy ambiguity, structural and reference ambiguities exist due to morphological richness of a language. To produce quality translation, it is necessary for a system to consider semantics of a text too. [30] enhanced ESAMPARK[17] by adding semantics in English to Urdu machine translation. Their semantically enriched knowledge based approach handled translation ambiguities using data mining and text mining techniques. [31] used moses for English to Urdu statistical machine translation. Their corpus consisted of 14,465 ahaadiths from Sahih Bukhari and Sahih Muslim.

---

Most of alignment was done manually and corpus was organized at hadith level rather than sentence level or word level. IRSTLM, MERT and BLEU were used for language modeling, tuning and evaluation of system respectively.

[32] discussed the structure and working of English to Urdu machine translation system. Their proposed system based on three modules: lexical module that converts input text into tokens, syntactical module that generates the parse structure and transformational module that converts English parse structure into Urdu. With this approach, they handled structural difference, word order in phrases and phrase order in sentences of both languages. [33] investigated word order issues in English-to-Urdu statistical machine translation. To solve this issue, they presented syntax-aware transformation based pre-processing technique which outperformed both lexical conditioned and distance based reordering models. The idea was to convert structure of source text (English) according to structure of target text (Urdu) at pre-processing level using transformation rules that are extracted from parallel corpus. [34] generated case markers in translating of text from fixed-order language (English) to free-order language (Urdu). Free order languages used case markers to highlight the relationship between the dependent noun and its head. For the purpose of uniform alignment in source and target sides, artificial case markers were introduced in source side (English) which improved BLEU score in phrase based machine translation and hierarchical phrase based machine translation as compared to baseline.

[35] applied factored model of statistical machine translation on English-Urdu language pair. Corpus was collected using java based web-crawler and pdf extraction programs. Factorization of target language (Urdu) was done at word level which helped to reduce data sparseness problems occurred due to lack of sufficient training data. BLEU score, METEOR, Precision, Recall and F-measure were used to evaluate the factored machine translation system which showed the improvements as compared to other statistical based machine translation systems for morphologically rich languages like Urdu. [36] used hierarchical phrase based model for English-Urdu machine translation. K-fold cross validation method was applied for sampling of corpus which were around 8000 sentences of EMILE corpus. *--glue-grammar* parameter and 3-gram language model were used with *maximum-phrase-length*=6. Due to language divergence of Urdu and English, classical phrase based worked better as compared to hierarchical phrase based. [37] presented machine translation approach which based on Case-based reasoning

(CBR) technique, translation rules base model and Artificial neural network work model. CBR solved future problems by using knowledge of similar past solved problems.

*Table II: Studies about Machine Translation*

| Study | Languages | Corpus | Technique |
|---|---|---|---|
| Le NT, Sadat F (2018) [38] | English-Vietnamese | News 2018 shared task | RNN |
| Mahata SK, Das D, Bandyopadhyay S (2018) [39] | English-Hindi | MTIL2017 shared task | SMT,RNN |
| Singh S, Anand Kumar M, Soman KP (2018) [40] | English-Punjabi | OPUS,TDIL,Crawled data | RNN with Encoder-Decoder |
| Ayesha MA, Noor S, Ramzan M, Khan HU, Shoaib M. (2017) [24] | Urdu-Arabic | Customized data | SMT, EBMT |
| Jawaid B, Kamran A, Bojar O. (2016) [34] | English-Urdu | UMC005 | PBSMT |
| Singh U, Goyal V, Lehal GS. (2016) [41] | Urdu-Punjabi | Manually Mapped | Incremental SMT |
| Durrani N, Koehn P. (2014) [42] | Urdu → Hindi Hindi→ English English→ Hindi | EMILE, Multi-indic | Phrase based SMT, Character based SMT |
| Malik AA, Habib A(2013) [28] | Urdu-English | Selected sentences | EBMT, RBMT, SMT |
| Khan N, Anwar MW, Bajwa UI, Durrani N (2013) [36] | English-Urdu | EMILE | Hierarchical PBMT |

| Durrani N, Sajjad H, Fraser A, Schmid H.(2010) | Hindi-Urdu | EMILE, Leipzig | Word based Translation, Character-based Transliteration |
|---|---|---|---|

## 2.2    Challenges in Machine Translation

There are three different types of challenges that researches may face in their work, neural machine translation challenges, general machine translation challenges, and Urdu, English structural challenges.

### 2.2.1   Neural Machine Translation Challenges

Neural machine translation has shown better results in recent research. Six type of challenges related to neural machine translation are discussed in [43]. In following section, that challenges are discussed.

**Domain Mismatch**

Different words have different translations in different domains. To produce domain specific or context aware translation is a big task in NMT as training data usually available in general perspective. For domain specific translation, different approaches can be applied, one of them is to train model on general training data first and then on domain specific data for some epochs.

**Amount of Data**

Amount of training data always plays an important role in model/algorithm training. Increasing the number of training instances will lead to better results. In NMT, training data should be in millions so that model can learn underlying pattern of data effectively.

**Out-of-Vocabulary Words**

To handle out-of-vocabulary (OOV) or rare words, is a challenging task in every machine translation system. Reasons of OOV or rare words includes word missing in training data, misspelled words, technical terms and foreign words that usually cannot be translated. Mina-Thang and Ilya [44] proposed a technique to handle rare words problem in NMT using word alignment model, emitting position of rare word in source sentence and then translating rare word using a dictionary. Their approach improves overall results up to 2.8 BLEU points.

**Long Sentences**

Another challenge is to maintain translation quality of the system on longer sentences. Kohen [43] empirically shows that NMT produces significant results on sentences having length upto 60 words. The performance of NMT decreases on sentences where length increases from 60 words which may be improved by introducing extra layers in network.

**Word Alignment**

Word alignment is another issue in neural machine translation. Usually, the attention based neural machine translation models does not play the role in word alignment from source to target sentences.

**Beam Search**

In neural machine translation, search strategy is the major issue for extracting best translation of given word. In NMT, this is done by simple beam search decoder which finds the best suitable translation by translating word by word having fixed size of beam. Having fixed size of beam, it gives good results for narrow and small beams only. To address this issue, [45] has discussed different pruning techniques.

### 2.2.2 General Machine Translation Challenges
**Word Translation Problem**

In each language, some words have different meanings in different context. To select appropriate translation according to situation is real problem in machine translation.

**Word Translation Problem**

Phrase translation is another challenge in machine translation. Word by word translation cannot be good option in proverb and idiomatic sentences.

### 2.2.3 Syntactic & Semantic Challenges
Urdu to English machine translation has some unique challenges which we have to overcome to produce reliable and context-aware translation system [32]. This section describes some of them by providing comparative analysis of syntactic and semantic differences of Urdu and English.

**Different Structures**

Urdu and English belongs to different family of structures. English is classified as SVO language because words in a sentence are present in order of Subject-Verb-Object. While Urdu is classified as SOV language because words in a sentence are present in order of Subject-Object-Verb. Following table represents example sentence of both languages.

| English | | | Urdu | | |
|---|---|---|---|---|---|
| I | eat | food | کھاتا ہوں | کھانا | میں |
| Subject | Verb | Object | Verb | Object | Subject |

**Word Order in Phrase**

Urdu and English languages also differs with respect to order of words in particular phrases. For example, in prepositional phrases the position of ***head word*** is different in both languages. Head word is described as "". In Urdu, head word is the last word of sentence while in English it is not. Following table describes the examples of both.

| English | | | Urdu | | |
|---|---|---|---|---|---|
| Flag | of | Pakistan | جھنڈا | کا | پاکستان |
| Noun(Head) | Preposition | Noun | Noun(Head) | Preposition | Noun |

**Different Forms of Single Word**

Usually, a single word can be represented in different forms according to formation of sentence and type of tense. For example, verb in English can be represented in five different forms. The word "eat" has following five forms: eat, eats, ate, eaten, eating and these forms will be used with respect to position of words and nature of tense. In Urdu, there are some more forms of words. Words transforms their form not only on the basis of tense but also on the basis of subject/noun type.

| English | Urdu |
|---|---|
| He **eats** food. | وہ کھانا کھاتا ہے |
| She **eats** food. | وہ کھانا کھاتی ہے |
| Green Flag | ہرا جھنڈا |
| Green Flags | ہرے جھنڈے |
| Green Chair | ہری کرسی |
| Green Chairs | ہری کرسیاں |

**Words having Gender**

In English, usually adjectives and nouns do not associated gender with them. But in Urdu, adjectives and nouns have gender associated with them for example, chair is considered as feminine and table is considered as masculine in Urdu.

| English | Urdu |
|---------|------|
| Chair | کرسی |
| Table | میز |

**Types of Tense**

Another difference between Urdu and English is the types and number of tenses. For example, past indefinite tense of English language can be further divided into three categories in Urdu language. Since there is no one-to-one mapping between tenses of both languages, so it be very challenging to provide accurate translation. Following table describes the examples:

| English | Urdu | |
|---------|------|---|
| He ate food. | اس نے کھانا کھایا ہے | (Absolute Past) |
| | اس نے کھانا کھایا | (Near Past) |
| | اس نے کھانا کھایا تھا | (Distant Past) |

**Identification of proper nouns and abbreviations**

English follows the rule of capitalizing the first character of proper noun and for abbreviation, all characters of word will be capitalized. For example, USA is an abbreviation of United States of America. Unlike English, there is no such rules for identification of proper nouns and abbreviation which makes very hard to classify the word either it is proper noun or noun. Following table describes the scenario.

| English | Urdu | Type of Word | Example sentence |
|---------|------|--------------|------------------|
| Great | عظیم | Noun | Quaid e Azam was a great leader. |
| Azeem | عظیم | Proper-noun | My name is Azeem. |
| PAK | پاک | Abbervation of Pakistan | Indo-Pak match is scheduled tomorrow. |
| Clean/Holy | پاک | Noun | This place is clean. |

# Chapter 3
# Methodology

Machine translation of Urdu language is still infancy and lacking basic translation approach which provides acceptable translation of Urdu text. The current state-of-art translation engines which provides good translation of many language pairs still struggling on Urdu language.

The proposed model of machine transliteration and translation consists of following steps:

Step I.    Corpus (tokenization, alignment)

Step II.    Transliteration

Step III.    Translation

## 3.1    Corpus (tokenization + alignment)

The amount of parallel corpus and its quality plays significant role in quality of translation output. For low resource languages like Urdu, it is extremely difficult to find sufficient parallel corpus for training, validation and testing of translation engine. For our experiment, we used Quran data of corpus UMC005[18]. The corpus consists of 6414 sentence pairs dividing into three categories training, validation and testing. The stats of corpus are given as:

**Training-set**

While applying **cleaning** script with sentence length maximum to 80, training sentences reduced from 6000 to 5419.

|  | Urdu | English |
|---|---|---|
| # of sentences | 5419 | |
| Avg. sentence length | 32 | 33 |
| Min , Max words in sentence | 5,80 | 4,80 |
| # of total words | 177797 | 183554 |
| # of unique words | 6246 | 7586 |

**Validation-set**

|  | Urdu | English |
|---|---|---|
| # of sentences | 214 | |
| Avg. sentence length | 15 | 16 |
| Min , Max words in sentence | 7, 56 | 4,43 |
| # of total words | 3375 | 3596 |
| # of unique words | 908 | 937 |

**Test-set**

|  | Urdu | English |
|---|---|---|
| # of sentences | 200 | |
| Avg. sentence length | 17 | 17 |
| Min , Max words in sentence | 5,67 | 2,65 |
| # of total words | 3571 | 3591 |
| # of unique words | 903 | 938 |

Tokenization of data is performed by freely available tokenizer.perl script of Moses toolkit[19]. For word-alignment GIZA++ [46] is used where alignment was symmetrized by *grow-by-diagonal* and *heuristic*.

## 3.2    Transliteration

Machine Translation models suffers from OOV (out-of-vocabulary) words which are mostly technical terms, loan words of other languages or the named entities. Transliteration has shown the improvements of MT [19] . The challenges part is the lack of transliteration corpus for most of the languages, and if data available, integration or use of transliterated words in training of machine translation engine is not available. Mostly a supervised transliteration module trained outside the MT engine where idea is to replace OOV words with their 1-best translation and then integrate into main system.  In our approach, there is no need of explicit training of transliteration module i.e approach is un-supervised, language independent and model learn from un-labelled parallel data. The work is based on transliteration mining adapted from the work of [47] [48]. The character alignment of source word $e$ and its corresponding word $f$ in parallel data can be found in many ways. In mathematical form, alignment sequence of source $e$ and target $f$ can be referred as $a$. Function $Align(e,f)$ returns all possible alignment

---

sequences **a** of word **e** and **f**. The joint probabilities of a word pair is sum of all alignments sequences as discussed in [49].

$$p_1(e, f) = \sum_{a \in Align(e,f)} p(a) \qquad (1)$$

Expectation maximization (EP) algorithm is used to learn alignments between transliteration pairs. EP maximizes the likelihood of training data. There are three methods in machine transliteration based on transliteration mining and EP [48].

**Method 1:**     This method only replaces OOV words from the output with 1-best transliteration without considering context which may lead incorrect translation. The accuracy of method depends on the efficiency of transliteration module.

**Method 2:**     This method provides n-best transliteration to monotonic decoder. In this method, LMOOV feature is used to count unknown words to language model and KneserKey smoothing is used to assign probabilities to unseen events which may lead to incorrect translation.

**Method 3:**     The shortcomings of method 1 and method 2 are addressed in this approach. In this method, phrase-table of transliteration module is fed to decoder for better reordering of unknown words. The option *decoding-graph-backoff* is used for generation of multiple phrase-tables and back-off models.

## 3.3    Translation

At this steps, we trained three models of statistical machine translations and two models of neural networks on default settings known as baseline and baseline + transliteration. The effectiveness of transliteration module is also tested to know about the impact across different approaches. The details of different models of statistical machine translation and neural machine translation adapted in this experiment are given below.

## 3.4    Statistical Machine Translation

Statistical machine translation is based on maximum likelihood or related criteria. For example, if there is source sentence S= $S_1, S_2, S_3 \ldots\ldots S_n$ which will be translated to target sentence T= $T_1$, $T_2$, $T_3$.... $T_n$. Considering the fact that one word may have multiple translations which all are acceptable and can be arranged in a sentence in many ways. Among all of them, selection of correct translation is tough task and for this, a number(probability) is assigned to every translation. The best translation will be picked using the following formula [50]:

$$e(f) = argmax\{Pr(e|f)\} \qquad ..... (i)$$

where argmax function represents generation of output words in target language. As the probability distribution function is unknown, a model P(e|f) will be developed to approximates argmax Pr(e|f). Statistical machine translation depends on following alignment, phrase extraction and phrase reordering.

**Alignment:** Alignment plays an important role in correspondence of words and phrases of source language to words and phrases of target language. It can be used for extraction of phrases, generation of phrase-tables and generation and correctness of hypothesis. The better alignment in words and phrases of parallel corpora will help to generate more accurate and quality output of machine translation system. In earlier years, all statistical models were word based [51] and alignment between words of source sentence and target sentence was mostly, one-to-one which can one-to-many or vice versa. For example, a source sentence **S** = ($S_1$, $S_2$, $S_3$......, $S_n$) having length **n** generate target sentence **T** = ($T_1$, $T_2$, $T_3$ ......., $T_n$) having length **m** and set of alignment links can be define as A = ($A_1$, $A_2$, $A_3$ ....... $A_n$) where **$A_1$** is the link between **$S_1$** and **$T_1$** and so on.

Many researchers have used different models of statistical machine translation in their research and results are quite convincing. We implemented statistical model using moses toolkit.

### 3.4.1  Phrase based statistical machine translation

In phrase based translation, basic unit of translation is phrase where phrases are multiple occurrences of words or segments[20]. During translation, text of source language is converted into phrases, each phrase is translated into target language and then phrases are reordered. Phrase based statistical machine translations showed improved results as compared to word-based [52].

---

[20] http://www.statmt.org/moses/manual/manual.pdf

### 3.4.2   Hierarchical phrase based statistical machine translation

Hierarchical phrase based technique known as *phrases that contains sub phrases* is based on weighted synchronous context free grammar that can learn from parallel text without any syntactic annotation. [53]. Hierarchical phrase based shows improved results over classical phrase based due to better reordering and generalization techniques.

### 3.4.3   Factored based statistical machine translation

Factored based machine translation model which is an extension of phrase based model enables linguistic information such as *part-of-speech tags, stems and lemmas*.

## 3.5   Neural Machine Translation

Neural machine translation gains much attention in recent years which is based on neural networks. Neural network is set of algorithms, loosely coupled to mimic the behaviour of human brain. It is a layered architecture where layers are made up of interconnected nodes which contains activation function. Typically, it has three types of layers, *input layer* which receives input and then communicate to *hidden layer* which can be of different numbers and usually responsible for all computation and then *hidden layer* connected to *output layer* where the actual output receives. Traditional statistical machine systems are made up of many sub-components that are tuned separately while the purpose of neural machine translation is to build a single, large neural network that reads a sentence and outputs its translation. Neural networks model the conditional probabilities **p(y|x)** of a source sentence $x_1$, $x_2$, $x_3$…, $x_n$ to a target sentence $y_1$, $y_2$, $y_3$…, $y_n$. A very basic neural network has two components **a)** an encoder which reads input sentence and encode the sentence into fixed-length vector representation **b)** a decoder outputs a translation from vector representation and decomposes conditional probability as [54]:

$$\log p(y|x) = \sum_{j=1}^{m} \log p\left(y_j | y_{<j}, \boldsymbol{s}\right)$$

Encoder-decoder system is jointly trained to maximize the probability of a translation provided the source sentence.

The fundamental purpose of neural networks is to train a model on some training-data having set of learnable features to make predictions on data which is inputted later. The data/corpora used in training of model consists of input $x$ having features $f_x$ and output $y$. Usually, the output $y$ is called class or label in simple machine learning tasks. In complex tasks of machine learning like machine translation, the output $y$ also consists of features $f_y$ just like input $x$. Any neural network follows following steps:

➢ Preprocessing: to remove undesired characters/data from training corpora or make it readable/executable for machine learning model.

➢ Usually, the corpora are divided into *training*, *validation* and *testing* with split-size of 70%, 20% and 10% respectively and sometime in *training* and *testing* with 80-20 rule.

➢ The training-split of corpora is divided further into mini chunks and fed to the model.

Usually any neural network follows four steps at very basic level. In first step, the model took input $x_i$ for **i** in $\mathbf{x \in [1, n]}$ and predict a label $y_i$.

$$\hat{y}_i = \text{softmax}(W x_i)$$

Where $W$ is referred as weight matrix whose dimensions are $(\mathbf{f_x}, \mathbf{f_y})$ features of input $x$ and output $\mathbf{y}$ and main responsibility is to predict label $y_i$ for input $x_i$. The multiplication of matrix $W$ and vector $x$ convert input-vector of length $f_x$ (number of input features) into output-vector of length $f_y$ (number of labels/classes). The probability of each class known as $c$ will be computed as:

$$P(c_j) = \frac{e^{c_j}}{\sum_{k=1}^{L_y} e^{c_k}} \quad \forall \text{ class } c_j \text{ where } j \in [1, L_y]$$

With these predicted classes, the second step of neural network is to compute the loss $l_i$. To measure the accuracy of predicted labels $\mathbf{\hat{y}}$, a number of different functions can be used in ***LossFunction*** as:

$$l_i = LossFunction(\hat{y}_i, y_i)$$

At third step of training, first and second steps are repeated for $x_i$ for every **i** in $\in \mathbf{[1, n]}$, to compute total loss $L$ through entire corpora/dataset in a single iteration called epoch.

$$L = \sum_{i=1}^{n} l_i$$

At fourth and final step of training, $\partial L/\partial W$ is used to update $W$ which is obtained by differentiating $L$ w.r.t $W$.

$$W = W - lr(\partial L/\partial W)$$

Where $lr$ is hypermeter referred as learning rate. These four steps are repeated iteratively for specific number of epochs. To determine the performance of the model, accuracy of the model is computed on validation test at some specific points (e.g., after every 5 epoch). This is done by computing **Loss** between step 1-3 just like training process. This validation process is very important as it help to determine the progress of model on unseen data outside the training data which help to avoid overfitting. The process described above is single layer neural network with one weight matrix, however in deep neural networks several weight matrices can be applied to input data. That four steps of neural network training are described in following figure [55]:

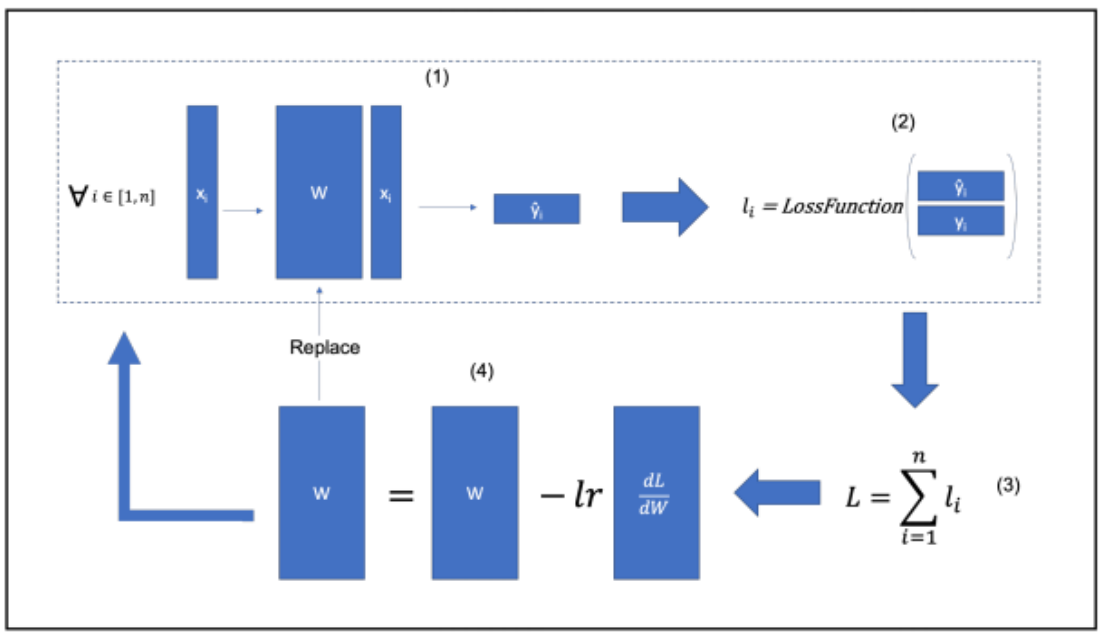

*Figure 2: Training process of one-layer neural network*

### 3.5.1   Recurrent Neural Network (RNN)

In single layer neural networks, information flows in one direction only. These type of networks are also called feed-forward since information flows in forward direction from input layer to output layer and there is no cycles or loops in them. For time series and textual data, neural network should be able to memorize the parts of input and uses that knowledge to make accurate predictions.

RNNs are specialized networks able to handle textual and time series data. In case of time series data, input of each time series point $x_i$, composed of a series of points $x_1$, $x_2$, $x_3$…., $x_n$ where n is the size of time series. The main objective of recurrent neural network (RNN), just like others, is to predict the label/class $y_i$ of input $x_i$. The training of recurrent neural networks is same like standard neural network with some tweaks at step 1 and 4.

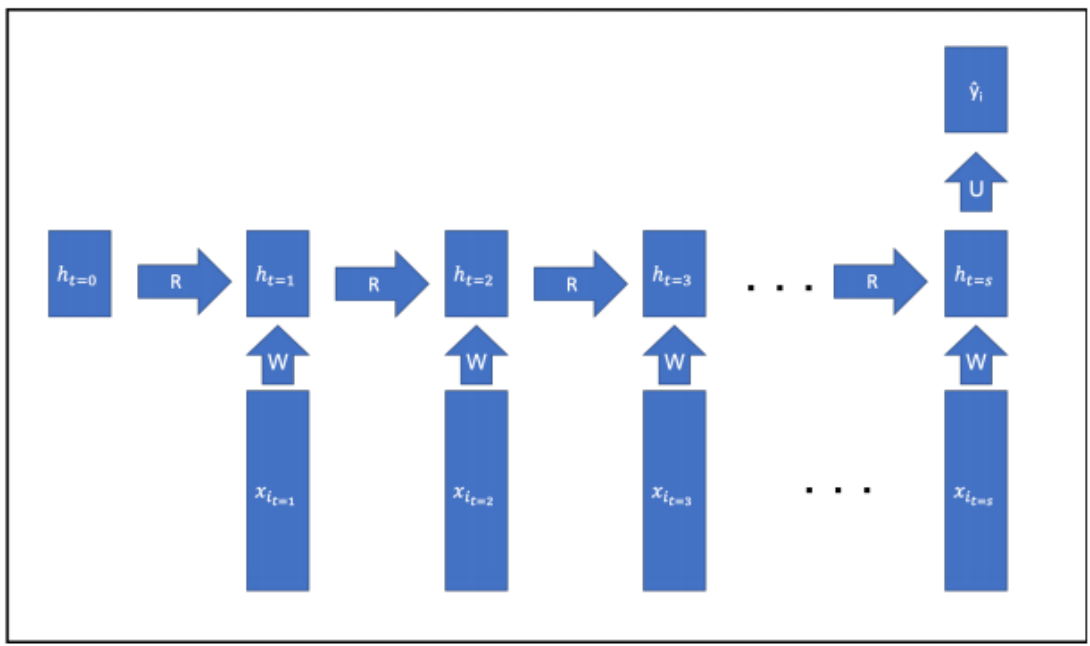

*Figure 3: Predicting a label/class using RNN*

In first step of RNN's training, it consecutively analyzes each point $x_i$ upto $x_n$ and stores the information in hidden state $h_t$ to make a prediction. At the end of each sequence, the $h_{t=0}$ is initialized to 0 and updated at each time step **t** using following formula:

$$h_t = \sigma(Rh_{t-1} + Wx_{i_t})$$

where **R, W** are weight matrix and **σ** is non-linear function also called sigmoid. By applying this architecture, at any time step **t,** the model stores the information at hidden state $h_t$ which can applied at to make a prediction. The step 2 & 3, **softmax** function, calculation of **LossFunction**, updation of weight matrixes and all the internal working of **rnn** training, are exactly same as discussed in section of standard neural network.

Among all the neural networks, models with Long-short term memory (LSTM) and attention mechanism (transformer) have proved their effectiveness in NMT.

### 3.5.2   Encoder-Decoder & Encoding of Input

The main objective of machine translation system is to translate source sentence of one language to target sentence of other language. Each word in dataset is converted into fixed-length vector representation by encoder and vocabulary lists are created for both input and output languages which contains all unique words along with <SOS> start of sentence and <EOS> end of sentence that are useful in training. First of all, the vocabulary is created from training corpora and textual data is converted into numeric form by encoder. For example, we have input sentence کراچی پاکستان کا ابم شہر اور تجارتی بندرگاہ ہے, its vocabulary table and encoding will be given as:

*Table III: Input sentence vocabulary table*

| 0 | شہر |
|---|---|
| 1 | پاکستان |
| 2 | کا |
| 3 | ابم |
| 4 | کراچی |
| 5 | اور |
| 6 | ہے |
| 7 | بندرگاہ |
| 8 | تجارتی |
| 9 | <SOS> |
| 10 | <EOS> |

$$\text{کراچی} = \begin{bmatrix} 0 \\ 1 \\ 0 \\ 0 \\ 0 \\ 0 \\ 0 \\ 0 \\ 0 \\ 0 \\ 0 \end{bmatrix} \quad \text{پاکستان} = \begin{bmatrix} 0 \\ 0 \\ 0 \\ 0 \\ 1 \\ 0 \\ 0 \\ 0 \\ 0 \\ 0 \\ 0 \end{bmatrix} \quad \text{کا} = \begin{bmatrix} 0 \\ 0 \\ 0 \\ 0 \\ 0 \\ 0 \\ 0 \\ 0 \\ 0 \\ 0 \\ 0 \end{bmatrix} \quad \text{ابم} = \begin{bmatrix} 0 \\ 0 \\ 0 \\ 1 \\ 0 \\ 0 \\ 0 \\ 0 \\ 0 \\ 0 \\ 0 \end{bmatrix} \quad \text{شہر} = \begin{bmatrix} 0 \\ 0 \\ 0 \\ 0 \\ 0 \\ 0 \\ 0 \\ 0 \\ 0 \\ 0 \\ 0 \end{bmatrix} \quad \text{اور} = \begin{bmatrix} 0 \\ 0 \\ 0 \\ 0 \\ 0 \\ 1 \\ 0 \\ 0 \\ 0 \\ 0 \\ 0 \end{bmatrix}$$

*Figure 4: Encoding of input sentence*

NMT utilizes the Encoder-Decoder architecture. In this structure, a first RNN (the encoder) analyzes the input sentence and passes its final hidden state, $h^E_{t=Lx}$, onto a second RNN (the decoder) to use as its first hidden state, $h^D_{t=0}$. If a sentence has a length 7, the encoder portion is given as:

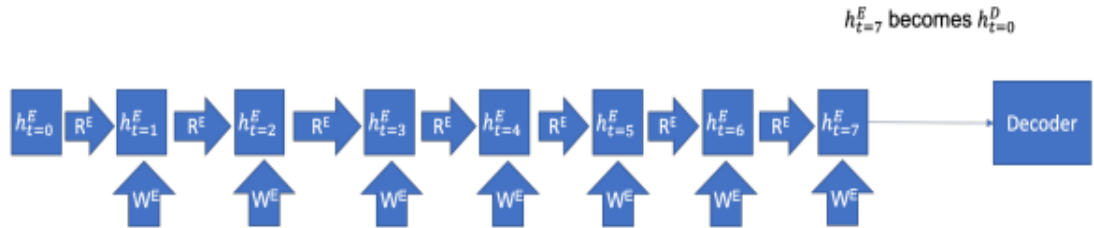

*Figure 5: Encoder portion of NMT*

Where $\mathbf{W}^E$ represents a single word at position $\mathbf{t}$i. And the decoder portion is given as:

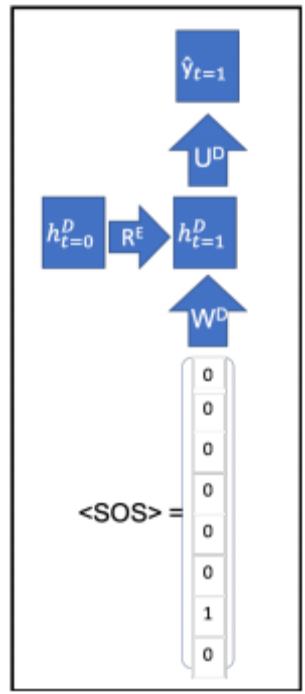

*Figure 6: Decoder portion of NMT*

### 3.5.3 Long-Short Term Memory Model

RNN suffers from short-term dependencies, if information is long enough it will struggle to preserve it. Long-short term memory (LSTM) is a specialized model of recurrent neural network (RNN) capable of learning long-term dependencies. It is able to remember inputs from up to 1000 time steps in the past. This capability makes LSTM an advantage for learning long sequences with long time lags.

The fundamental building blocks of LSTMs are cells or cell state which provide a bit of memory to LSTM so that it can remember. There are three gates *input-gate, forget-gate* and *output-gate* in LSTMs[21] which are sigmoid activation function which gives output between 0 and 1. LSTM calculates hidden state $h_t$ using following mathematical formulas:

$$i_t = \sigma\big(x_t U^i + h_{t-1} W^i\big)$$
$$f_t = \sigma\big(x_t U^f + h_{t-1} W^f\big)$$
$$o_t = \sigma\big(x_t U^o + h_{t-1} W^o\big)$$
$$\tilde{C}_t = \tanh\big(x_t U^g + h_{t-1} W^g\big)$$
$$C_t = \sigma\big(f_t * C_{t-1} + i_t * \tilde{C}_t\big)$$
$$h_t = \tanh(C_t) * o_t$$

Where *i, f* and *o* are input, forget and output gates respectively with different kind of parameters. *W* is recurrent connection between previous hidden layer and current hidden layer and *U* is weighted matrix responsible to connect input nodes to current hidden layer.

### 3.5.4 Transformer Model

The core idea behind transformer model is *self-attention*. LSTMs have some issues parallelization, long and short range dependencies and distance between positions is linear. To solve these problems, attention mechanism is introduced in neural networks. Each word has hidden state which is passed along the way while translating the sentence instead of decoding whole sentence in a single hidden unit/layer. To solve problem of parallelization, transformers used convolutional neural networks (CNN) along with attention mechanism. The mathematical form attention is given as:

$$Attention\big(Q, K, V\big) = softmax\left(\frac{QK^T}{\sqrt{d_k}}\right)V$$

---

[21] https://medium.com/@divyanshu132/lstm-and-its-equations-5ee9246d04af

Transformer uses attention mechanism in three ways [56]:

➢ In "encoder-decoder attention" layers, the queries come from the previous decoder layer, and the memory keys and values come from the output of the encoder. This allows every position in the decoder to attend over all positions in the input sequence. This mimics the typical encoder-decoder attention mechanisms in sequence-to-sequence models.

➢ The encoder contains self-attention layers. In a self-attention layer all of the keys, values and queries come from the same place, in this case, the output of the previous layer in the encoder. Each position in the encoder can attend to all positions in the previous layer of the encoder. Similarly, self-attention layers in the decoder allow each position in the decoder to attend to all positions in the decoder up to and including that position. We need to prevent leftward information flow in the decoder to preserve the auto-regressive property. We implement this inside of scaled dot-product attention by masking out (setting to $-\infty$) all values in the input of the softmax which correspond to illegal connections.

We implemented above models of NMT using OpenNMT-py[22] toolkit. For LSTM, we used default settings while for transformer model followings were parameter settings. Selection of these parameters helped researchers to mimic behaviour of google translator as reported in WNMT-18 [57]. To run these experiments, GPU Tesla k40 is used with 32 gb ram and graphic card of 12 gb. For LSTM model, it took around 10-12 hours to train model and for transformer model it took around 40 hours due to large number of hidden layers and rnn size.

*Table IV: NMT parameter selection*

| layers = 6 | batch_size = 4096 | learning_rate = 2 | max_grad_norm = 0 |
|---|---|---|---|
| rnn_size = 512 | batch_type = tokens | label_smoothing = 0.1 | param_init = 0 |
| word_vec_size= 512 | dropout = 0.1 | encoder_type = transformer | param_init_glorot |

---

[22] https://github.com/OpenNMT/OpenNMT-py

**Chapter 4**

**Results & Discussion**

## 4.1 Evaluation Measures

We have used following evaluation measures for checking the effectiveness of our approach:

- BLEU Score
- METEOR
- TER
- Precision
- Recall
- F1_Measure

### 4.1.1 BLEU Score

Bilingual evaluation understudy (BLEU) is one of fundamental evaluation measure in machine translation domain. It uses a modified form of precision recall to match output text against multiple reference sentences. The primary programing in BLEU implementer is to match the n-grams of candidate with n-grams of reference without considering the position of word. Mathematical formula of bleu score is:

$$P = M_{max} / W_t$$

**Output sentence**: In two weeks Pakistan's weapons will give army.

**Reference 1**: The Pakistani weapons are to be handed over to the army within two weeks.

**Reference 2**: The Pakistani weapons will be surrendered to the army in two weeks.

6/8 = BLEU score is 0.75

Where 8 is total length of output sentence and 6 tokens are matched with references.

### 4.1.2 METEOR

Metric for evaluation of translation with explicit reordering based on unigram-precision and unigram-recall is intended to improve BLEU score. METEOR is based on unigram matching between machine translation and human translation. This matching is based on surface-form, stem-form and meanings of unigram which can be extended further to more complex matching strategies. METEOR has the following formula as discussed in [58]:

$$\text{METEOR} = \frac{(10\ P\ R)}{(R + 9\ P)}\ (1 - Pm)$$

Where P is the unigram precision and R is the unigram recall. The MTEOR brevity $P_m$ is:

$$Pm = 0.5\ (\frac{C}{Mu})$$

### 4.1.3  TER

Translation error rate (TER) is one of newest edition in family of evaluation measures for machine translation. It represents the minimum edits required to change machine output so that it exactly matches the reference translation. These edits might include deleting, inserting and substituting of word or even phrases. Mathematical form of TER is as following [59]:

$$\text{TER} = \frac{\#\ \text{of edits}}{\text{average}\ \#\ \text{of reference words}}$$

### 4.1.4  Precision

Precision is the ratio of retrieved records that are related to the query   It is calculated by calculating by using the following formula:

$$\frac{|\{relevant\ records\} \cap \{retrieved\ records\}|}{|\{retrieved\ records\}|}$$

### 4.1.5  Recall

Recall is the ratio of related records that are retrieved successful. It is calculated by using the following formula:

$$\frac{|\{relevant\ records\} \cap \{retrieved\ records\}|}{relevant\ records}$$

Precision and recall are inversely proportion:

As Precision increases ↑ recall↓ decreases

Conversely

As Precision decreases ↓ recall increases ↑

### 4.1.6   F1_Meausre

F1 measure is calculated by calculating the precision and recall. It is basically harmonic mean of precision and recall. We used following formula for calculating the F1_Score:

$$2.\frac{Precision.recall}{Precision+recall}$$

## 4.2   Experimental Results

Our transliteration technique improved baseline score upto 0.63 to 0.91 in terms of BLEU score. We have applied our technique in different models of statistical machine translation to show the effectiveness of the technique.

*Table V: SMT Results*

| Evaluation Measures | Phrase Based (PB) Model | | Hierarchical PB Model | | Factored Model | |
|---|---|---|---|---|---|---|
| | Baseline | With transliteration | Baseline | With transliteration | Baseline | With transliteration |
| BLEU | 14.45 | 15.21 | 14.50 | 15.41 | 10.36 | 10.99 |
| METEOR | 16.5 | 20.4 | 20.8 | 20.5 | 18.9 | 18.6 |
| TER | 79.1 | 75.6 | 79.1 | 77.0 | 81.5 | 80.8 |
| Precision | 0.39 | 0.46 | 0.42 | 0.44 | 0.49 | 0.44 |
| Recall | 0.36 | 0.46 | 0.47 | 0.45 | 0.43 | 0.45 |
| F1 | 0.37 | 0.45 | 0.44 | 0.45 | 0.44 | 0.44 |

*Table VI: NMT Results*

| Evaluation Measures | LSTM Model | | Transformer Model | |
|---|---|---|---|---|
| | Baseline | With transliteration | Baseline | With transliteration |
| BLEU | 6.65 | 7.93 | 7.57 | 9.62 |
| METEOR | 16.5 | 20.4 | 15.7 | 21.3 |
| TER | 79.1 | 75.6 | 74.7 | 72.1 |
| Precision | 0.35 | 0.47 | 0.41 | 0.60 |
| Recall | 0.31 | 0.46 | 0.35 | 0.53 |
| F1 | 0.33 | 0.44 | 0.38 | 0.52 |

## 4.3 Discussion

Experiments are conducted to measure the effectiveness of transliteration technique on different machine translation models. In first experiment, we applied three statistical models of machine translation i.e phrase based, hierarchical phrase based and factor based models. Results showed the performance w.r.t different evaluation measures.

**Source-sentence:** قرابت دار یتیم کو ۔

**Reference:** **The orphaned relative.**

| Factor based | Phrase based PB | Hierarchical (PB) |
|---|---|---|
| the orphans kindred , ) . | the kindred , orphans ) | to the orphan ) . |

The output of translation system largely depends upon the quality of parallel corpus. Being low resource language, Urdu language don't have enough resources. The corpora used in training/validation of system contains too much punctuation marks which is reflecting in machine's output of above table. Most of evolution measures, checks the performance of any algorithm by applying different matching criteria which is between reference (human) translation and machine translation. Word *kindred* and *relative* in reference and output sentences are of same meanings. Different people might translate one sentence differently using different words (synonyms) or used different order/arrangements of words in sentence. Reference translation can also be effected by biasness of its translator. So in these particular scenarios, relying on only one reference translation and not considering all these factors may affect the performance of translation engine.

Urdu has different set of punctuation marks lies in category of Unicode characters. English has character **","** while in Urdu its equivalent is **"،"**, English language has following end-of-sentence or full-stop character **"."** while in Urdu end-of-sentence symbol is **"۔"**. In this way there is difference in, semicolon of English **";"** and semicolon of Urdu **"؛"**, single quote of English **"'"** and single quote of Urdu **"'"**, double quotes of English **"""** and double quotes of Urdu **""""**and question-mark of English **"?"** and question-mark of Urdu language **"؟"** sentence as compared to English. If these punctuations are not handled properly in pre-process step, they effect the working and quality of machine translation system. We performed experiments on raw and pre-processed corpora (where all these punctuations were handled) and there is gain of +3.5 BLEU score for processed corpora as compared to experiments done on raw corpora.

**Chapter 5**

**Conclusion**

In this work, transliteration approach is introduced in neural machine translation to improve baseline results. The approach discussed in this work is unsupervised and language independent. The efficiency of machine translation systems highly effected by out-of-vocabulary (OOV) words which include technical terms, foreign words and unknown words. The developed system learns the pattern of unknown words from its own using training corpus and it can be applied to any language pair. The proposed approach tested on five different machine translation techniques using six evaluation measures of natural language processing. On statistical machine translation techniques, we achieved the total difference of upto 0.63 to 0.91 in terms of BLEU score as compared to previous score which was 0.24 to 0.60. During our experiments, we observed that working of machine translation systems effected by a lot of factors and choosing the right combination of parameters/techniques may lead to better results. On neural machine translation systems, transformer model outperforms than LSTM model with score 7.75 to 6.65 and 11.61 to 9.08 on different experiments. Due to different structures, writing scripts and pronunciation marks of Urdu langauge w.r.t English language, selection of right preprocessing technique may lead to better results [60]. In order to assess the impact of tokenization, we experimented with raw corpus and with tokenized corpus and results showed improvements of +3.5 points in baseline BLEU score.

In future work, we would like to explore the preprocessing techniques in context of machine translation which will equally beneficial for Urdu language. During this work, we faced difficulties in order to train translation engine on large parallel corpus of Urdu-English language pair which is unfortunately not available yet. The creation of state-of-art parallel corpus for Urdu language can be interesting task to work on. Since machine translation systems largely rely on parallel corpus, un-supervised machine translation is new trend to bridge gap of un-availability of parallel corpus for low resource languages like Urdu.

**Chapter 6**

(PBSMT) and Neural Machine Translation (NMT) Systems," *Language in India,* vol. 18, no. 10, 2018.